\documentclass[lettersize,journal,twoside]{IEEEtran}

\usepackage{graphics}
\usepackage{bbding}
\usepackage{epsfig}
\usepackage{kpfonts}
\usepackage{times} 
\usepackage{amsfonts}
\usepackage{amsmath} % assumes amsmath package installed
\usepackage{amssymb}  % assumes amsmath package installed
\usepackage{mathrsfs}
\usepackage{algorithmic}
\usepackage{array}
\usepackage{epsfig} 

\usepackage{textcomp}
\usepackage{stfloats}
\usepackage{url}
\usepackage{verbatim}
\usepackage{xspace}
\usepackage{cite}

\makeatletter
\let\NAT@parse\undefined

\usepackage{subfigure}
\usepackage{booktabs} 
\usepackage{makecell}
\usepackage{hyperref}
\hypersetup{
    colorlinks=true,
    linkcolor=black,
    filecolor=magenta,    
    citecolor=black,
    urlcolor=blue,
    }
\usepackage[capitalize]{cleveref}
\usepackage{todonotes}
\usepackage{marginnote}
\presetkeys{todonotes}{size=\tiny}{}
\usepackage{caption}
\usepackage{caption}
\newcommand{\bigtimes}{\varprod}

\def\BibTeX{{\rm B\kern-.05em{\sc i\kern-.025em b}\kern-.08em
    T\kern-.1667em\lower.7ex\hbox{E}\kern-.125emX}}
\usepackage{balance}

\begin{document}
\title{\LARGE \bf Salient Sparse Visual Odometry with Pose-only Supervision}
\author{Siyu Chen$^{1}$, Kangcheng Liu$^{1}$,~\IEEEmembership{Member, IEEE}, 
Chen Wang$^{2}$, \\Shenghai Yuan$^{1}$,~\IEEEmembership{Member, IEEE}, 
Jianfei Yang$^{1 *}$, and Lihua Xie$^{1}$,~\IEEEmembership{Fellow, IEEE}
\vspace{-20pt}
\thanks{Manuscript received: Dec 9, 2023; Revised: Mar 9, 2024; Accepted: Mar 11, 2024. This paper was recommended for publication by Editor Sven Behnke upon valuation of the Associate Editor and Reviewers’ comments. }
\thanks{$^{\dagger}$ This work is supported by any National Research Foundation, Singapore under its Medium Sized Center for Advanced Robotics Technology Innovation}
\thanks{$^{1}$Siyu Chen, Kangcheng Liu, Shenghai Yuan, Jianfei Yang, and Lihua Xie are with the School of Electrical and Electronic Engineering, Nanyang Technological University, 50 Nanyang Avenue, Singapore 639798, {\tt\small \{siyu010, kangcheng.liu, yang0478, shyuan, elhxie\}@ntu.edu.sg}}%
\thanks{$^{2}$Chen Wang is with the Spatial AI \& Robotics Lab at The Department of Computer Science \& Engineering, State University of New York at Buffalo, Buffalo, NY 14260, USA, {\tt\small chenw@sairlab.org}}
\thanks{$^{*}$ Jianfei Yang is the team lead.}
\thanks{Digital Object Identifier (DOI):  see top of this page.}
}

\maketitle
\begin{abstract}
Visual Odometry (VO) is vital for the navigation of autonomous systems, providing accurate position and orientation estimates at reasonable costs. While traditional VO methods excel in some conditions, they struggle with challenges like variable lighting and motion blur. Deep learning-based VO, though more adaptable, can face generalization problems in new environments. Addressing these drawbacks, this paper presents a novel hybrid visual odometry (VO) framework that leverages pose-only supervision, offering a balanced solution between robustness and the need for extensive labeling. We propose two cost-effective and innovative designs: a self-supervised homographic pre-training for enhancing optical flow learning from pose-only labels and a random patch-based salient point detection strategy for more accurate optical flow patch extraction. These designs eliminate the need for dense optical flow labels for training and significantly improve the generalization capability of the system in diverse and challenging environments. Our pose-only supervised method achieves competitive performance on standard datasets and greater robustness and generalization ability in extreme and unseen scenarios, even compared to dense optical flow-supervised state-of-the-art methods.

\end{abstract}

% \begin{IEEEkeywords}
% Class, IEEEtran, \LaTeX, paper, style, template, typesetting.
% \end{IEEEkeywords}

\section{Introduction}
Visual odometry (VO) is a fundamental technique in enabling autonomous systems to estimate position and orientation with respect to their surrounding environment. It has wide applications in autonomous driving, AR/VR, and mobile robotics due to its inexpensive nature and rich sensory information~\cite{vslam}. Classical geometry-based methods~\cite{dso,orbslam3,svo, ral1, ral2, mur2017visual} perform well under favorable conditions but struggle in challenging situations marked by changes in illumination, motion blur, and sensor noises~\cite{deepvo}. Such limitations show the lack of robustness for traditional VO. In response, end-to-end deep learning-based techniques~\cite{deepvo, tartanvo, deepmemory, selfvo, undeepvo, ral_ccvo} are proposed to address these challenges but suffer from reduced performance when tested under conditions different from training data~\cite{dfvo}. 

To address the limitations present in various techniques, a new hybrid VO framework has been introduced~\cite{droidslam, dpvo, dfvo, banet}. This framework utilizes networks to pinpoint point correspondences while retaining geometric optimization to determine the poses. Broadly, there are two types of hybrid methods: dense methods, which identify the correspondence for each pixel, and sparse methods, which determine the correspondence for only a selected number of pixels to enhance efficiency. In contrast, many of the current methods either employ a densely supervised technique, demanding a large number of labels, or an unsupervised one, which cannot guarantee consistent results. Supervised techniques necessitate accurate optical flow (or depth) and camera positions. Yet, procuring accurate optical flow is notably tough in many real-world datasets~\cite{flowhard}. On the other hand, unsupervised techniques do not require annotations but are largely reliant on factors like photometric consistency and the assumption of image sharpness. 
Moreover, models developed with these assumptions are prone to challenges like motion blur and varying light conditions.

Pose-only supervision emerges as a viable strategy to balance model accuracy, robustness in complex scenarios, and labeling costs for sparse hybrid methods, partly because pose ground truth can be obtained easily in the real world through various methods \cite{nguyen2022ntu,mcdviral2024}. 
However, achieving good performance with pose-only supervision presents challenges. The lack of ground-truth pixel-level correspondence makes optical flow supervision complicated, leading to ambiguities in flow estimations. Furthermore, sparse hybrid VO methods \cite{dpvo}, which employ random tracking reference selection, simplify the process but create another layer of ambiguities in tracking. These layers of ambiguities are problematic in various common scenes and diminish the robustness, especially in conditions like overexposure or significant lighting variations.

This paper discovers that two extremely simple designs can eliminate the reliance on dense optical flow labels and boost generalization in new environments, resulting in a pose-only supervision model.
First, we find that a self-supervised homographic pre-training phase can substantially improve the network's learning of optical flow from pose labels, providing directional guidance.
Second, we find that replacing the strategy of optical flow patch extraction from random selection to salient selection could result in more robust pose estimation.
In the experiments, we show that these two simplified techniques achieve comparable performance on established benchmarks, and significantly enhance generalization, outperforming the state-of-the-art methods in real-world tests.

Our contributions can be summarized as:
\begin{itemize}
    \item  To the best of our knowledge, we are the first to investigate the hybrid sparse optical flow-based Visual Odometry with pose-only supervision. We unveil a groundbreaking self-supervised homographic pre-training method for optical flow. 
    This approach empowers the network to refine its optical flow estimation capabilities and bolster feature representations from just one image, proving advantageous for the subsequent sparse optical flow-based VO tasks that depend exclusively on pose supervision.
    \item A salient patch detection module and a salient patch refining step are introduced in the proposed system. The salient point detection module identifies those points with significant image features, striving to retain valuable patches while discarding unnecessary ones, and the salient patch refining training step enhances the network's cooperation with salient patches, thus improving accuracy and reliability, particularly in monotonous environments.
    \item Extensive experiments show that our pose-only supervised method achieves competitive results on one public dataset, better results on three public divergent datasets, and much greater robustness and generalization in extreme and unseen scenarios even compared with the state-of-the-art methods with dense optical flow supervision.
\end{itemize}

\begin{figure*}[t]
\centering 
\includegraphics[width=1\textwidth]{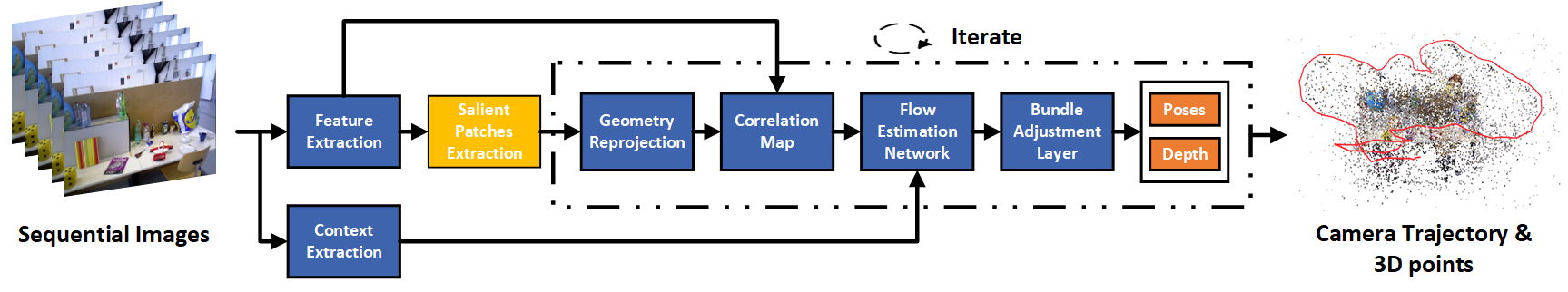}
\caption{Structure of our proposed method. Our method employs a CNN to extract features and patches from the salient patches extraction module. These patches are reprojected to neighboring frames by using the estimated poses and depth, and the correlation map is computed with the neighbor features of the reprojected positions. The correlation map, along with patch context information, is fed into the flow estimation network to get the optical flow and confidence weights. Then, the weighted bundle adjustment layer is applied to get poses and patch depths. This sequence—reprojection, correlation map computation, flow estimation, and bundle adjustment—is iterated N times to get the final poses and depth.}
\label{structure}
\end{figure*}

\section{Related Works}
% In this section, we introduce the learning-based VO methods, especially their corresponding supervision modes. 
In this section, we delve into learning-based VO methodologies, emphasizing their respective supervisory paradigms.

\textbf{Optical Flow and Pose Supervision} 
% Hybrid methods frequently rely on dense supervision for enhanced results. 
TartanVO~\cite{tartanvo} trains an optical flow network and then uses a CNN model to estimate poses from it.
Droid-SLAM~\cite{droidslam} estimates dense optical flows and the flow covariance, subsequently applying weighted bundle adjustment to deduce poses and depths. 
% This method uses the optical flow and pose labels for supervision. 
% Dense supervision is commonly used in hybrid methods for superior performances. Droid-SLAM~\cite{droidslam} uses a network to iteratively estimate the dense optical flows and their associated covariance and then uses weighted bundle adjustment to get poses and depths, and the labels of optical flow and poses are token as supervision. 
DPVO~\cite{dpvo} introduced a sparse optical flow-based approach, reducing computational costs by randomly selecting sparse patches for optical flow estimations rather than dense optical flow. 
Nonetheless, it still necessitates strong supervision with optical flow and poses. 
Moreover, its random patch selection approach, which introduces unreliability in inference and results in redundant computations. 
Compared with end-to-end methods, the hybrid methods have better performance and generalizability~\cite{dfvo, liu2022new}. 
However, dense supervision, especially optical flow, is difficult to carry out in real-world scenarios which limits the application values.

\textbf{Unsupervised Methods}
Several unsupervised methods have been developed for both end-to-end and hybrid systems to address the constraints of labeled data.
% Various unsupervised approaches have been proposed for both end-to-end methods and hybrid methods, with the goal of overcoming the limitations of labeled data. 
Li \textit{et al.}~\cite{undeepvo} and De Maio \textit{et al.}~\cite{de2020simultaneously} respectively propose an unsupervised end-to-end method, building upon DeepVO~\cite{deepvo} which relies on the photometric assumption.  DeFlowSLAM~\cite{deflowslam} introduced an unsupervised hybrid method based on photometric consistency between flow and poses.
% It deserves to be noted that most of the unsupervised methods are dependent on the assumptions of the photometric consistency and the sharpness of the images which is also hard for many datasets to satisfy and leads to the trained module still being sensitive to the motion blur, or illumination changing environments and therefore, restrictive. 
It's crucial to note that the majority of unsupervised methodologies hinge heavily on photometric consistency and image clarity assumptions. These can be challenging for many datasets, making the models sensitive to motion blur or varied lighting conditions.

\textbf{Pose Supersivion} 
% Pose supervision is mainly applied to end-to-end methods for camera pose estimations. 
DeepVO~\cite{deepvo} pioneers predicting relative camera pose by employing a CNN for feature extraction and subsequently integrating past information via a ConvLSTM.
% but has a bad performance in practice. 
However, such an end-to-end method is very likely to fail if the scenes are distinct from the training datasets~\cite{dfvo}.
To the best of our knowledge, the only hybrid approach is a structure-from-motion (SFM) technique~\cite{backtofeature} that applies a bundle adjustment layer on feature maps for pose estimation, supervised by a robust loss. However, due to high computational demands, this method is not suitable for real-time VO tasks.
% To the best of our knowledge, only a structure-from-motion (SFM) method~\cite{backtofeature}, which utilizes the BA layer directly on the feature map for pose estimation, can be seen as a hybrid method and is supervised by the poses with the help of a robust loss, but as an SFM method,~\cite{backtofeature} is not applicable for real-time VO task for the computational complexity.
In summary, pose-supervised, optical flow-based hybrid VO methods remain unexplored.
% In summary, methods with pose-only supervision in the category of optical flow-based hybrid VO remain unexplored.

\section{Approach}
As depicted in \cref{structure}, our hybrid VO system integrates two key elements: a salient random patch flow estimation module utilizing a network for flow estimation to enhance robustness and a weighted bundle adjustment layer employing the optimization layer for poses and depths.
The framework is inspired by~\cite{dpvo} but eliminates the need for dense optical flow supervision.
During the training phase, given that information on optical flow estimation is unavailable, we employ self-supervised homography pre-training, enabling the network to assimilate motion data.
Finally, the refinement of salient patches after standard pose-only training, to enhance their synergy with salient patches, is adopted. 
% After concluding the standard pose-only supervision training, we refine the salient patches to bolster their interaction with the salient patches. 
% We next present the details of the structure components and the training process.

\subsection{Salient Patches-based Sparse Flow Estimation}\label{sec31}
As highlighted in~\cite{islam}, areas with images lacking texture are more likely to introduce noise for flow estimations. Particularly, in real-world applications, camera overexposure, and motion blur frequently occur and diminish the texture of the environment. 
In contrast to dense optical flow methods~\cite{droidslam, deflowslam}, which estimate optical flow for each pixel, our approach focuses on sparse estimation, which selects parts of the optical flow. We integrate salient part detection into the system to enhance robustness and stability, building upon the principles in~\cite{dpvo}. As shown in ~\cref{structure}, we first extract the feature map of the $i$-th frame using a convolution neural network represented as $F^i$ and then identify salient patches from this feature map for sparse flow estimation. Semi-direct methods~\cite{svo} often adopt FAST~\cite{FAST} to select the pixels whose intensity is very different from the neighbors as the center of the patches for tracking. Inspired by this idea, the points whose features are very distinct from the neighbors as the center of the patches are detected for efficient tracking. 
From~\cite{d2net}, the calculation of salient scores is divided into two parts, layer salient score $\alpha_{m,n,h}$ and channel salient $\beta_{m,n,h}$ scores, shown as follows:
\begin{equation}
\begin{aligned}
\alpha_{m,n,h} = \frac{\exp(F_{m,n,h})}{\sum_{(\tilde{m},\tilde{n}) \in \mathcal{N}(m,n)} \exp(F_{\tilde{m},\tilde{n},h})}
\end{aligned},
\end{equation}
and 
\begin{equation}
\begin{aligned}
\beta_{m,n,h} = \frac{F_{m,n,h}}{\max_{g}(F_{m,n,g})},
\end{aligned}
\end{equation} 
where $F_{m,n,h}$ denotes the element of the $m$-th row, the $n$-th column, and the $h$-th channel of the feature map, and $\mathcal{N}(m,n)$ denote the neighbor of the $m$-th row and the $n$-th columns pixel. Then the final score of the salience can be presented as:
\begin{equation}
\begin{aligned}
s_{m,n} = \max_{h}(\alpha_{m,n,h} \cdot \beta_{m,n,h}).
\end{aligned}
\end{equation} 
After the salient scores are obtained, we grid the pixels of the frames and then get the highest score in each grid. Non-maximum suppression (NMS) is then used to eliminate redundant points and then the top $k$ patch with highest scores are selected. The $k$-th patch on the $i$-th frame can be expressed as $P_k^i = [x+w_1,y+w_2,1,d]^{\scriptscriptstyle T} \in \mathbb{R}^{4 \times (2r+1)^2}, w_1, w_2 \in \mathbb{Z} \cap [-r,r]$, where $(x,y)$, $d$ and $r$ are the center position, the depth, and the radius of the patch, respectively.
% With the salient patches detection module, the patches are selected as much as possible in informative places like the edge of the objects and add merits of the robustness of drastic motion and significant luminosity changes for the system.

Then, following~\cite{dpvo}, the estimated flows and the corresponding confidence weights can be obtained by performing three steps in order, the geometry reprojection, the correlation, and the flow estimation. The reprojection module is to reproject the patches on the $i$-th frame to its neighbor frames via the relative poses and the depths of the patches, and the reprojected position of the $k$-th patch from the $i$-th frame to the $j$-th frame can be expressed as $\tilde{P}_{k}^{ij} \sim K(T^j) (T^i)^{-1}K^{-1}P_k^i$, where $T^i \in \mathbb{SE}(3) \subset R^{4 \times 4}$ means the pose of $i$-th frame, and $K = [\mathcal{K}, 0;0, 1] \in R^{4 \times 4}$ is the projection matrix where $\mathcal{K} \in R^{3 \times 3}$ is the intrinsic matrix of the camera.
The correlation map $C$ can be calculated as  $C^{ijk}_{uv\mu \nu} = <F^j[\tilde{P}^{ij}_{k} + \Delta_{u v}], F^i[P^i_k]_{\mu \nu}>$ where $[\cdot]$ is the look-up operator, $<\cdot, \cdot>$ is the dot product operator, $\Delta_{u v}$ is $s \times s$ integer grid centered at $0$ indexed by $u$ and $v$, and $\mu \nu$ is the index of the coordinates of the patch pixels, respectively. A flow estimation network $\Psi(\cdot)$ is used to get updated flows $\delta$ and their confidence weights $\Sigma$, expressed as $\delta, \Sigma = \Psi(C, \mathscr{C})$ where $\mathscr{C}$ is the context information extracted from the source frame by a CNN. Then, the estimated patch position can be written as $\hat{P}^{ij}_k = \tilde{P}^{ij}_k + \delta^{ij}_k$. 

\begin{figure}[t]
\centering 
\includegraphics[width=1\linewidth]{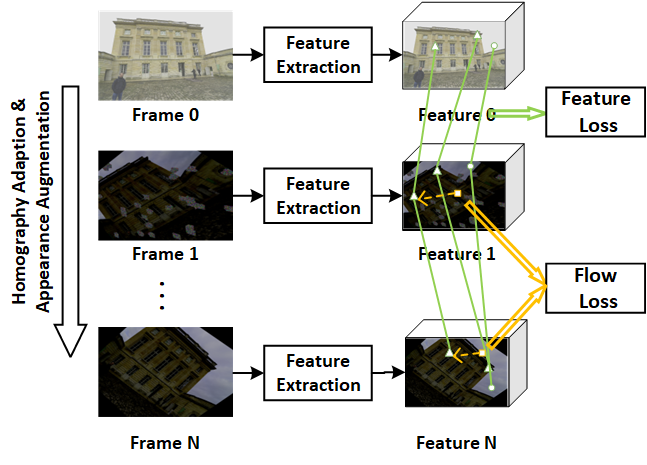}
\caption{The illustration of the self-supervised training process. The green triangle and circles denote the salient patches and the randomly selected patches, respectively. The yellow squares denote the estimated flow. The corresponding points are obtained by homographic adaption as the ground truth for the flow training and the feature training.}
\label{selftrain} 
\end{figure}

\begin{figure*}[ht]
\centering 
\includegraphics[width=\textwidth]{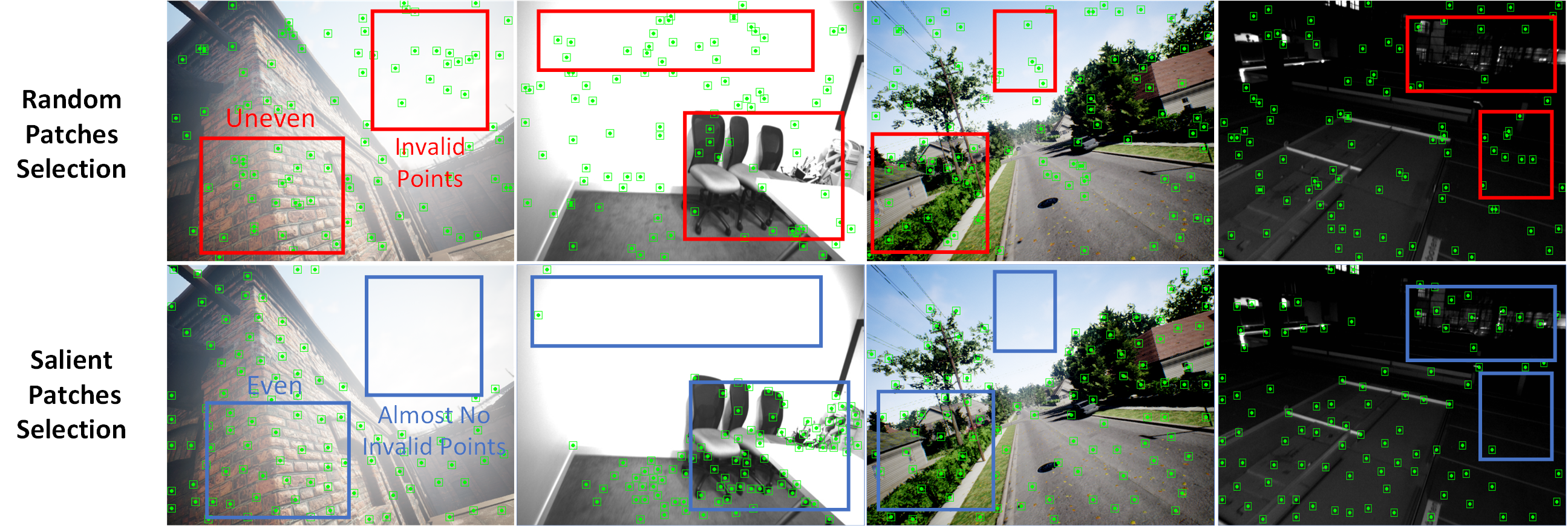}
\caption{The comparisons of the random selection patches. The green square means the selected patches. The first line shows the random patches selection strategy of DPVO~\cite{dpvo} and the second line shows the salient patches selection strategy of our method. Our method can provide more meaningful and even patches compared with the random selection strategy.}
\label{kpcompare} 
\end{figure*}

\subsection{Self-Supervised Homography Pre-Training} \label{sec32}
Due to the absence of the optical flow ground truth during the training process, the network relies solely on learning optical flow information from the available pose labels. 
The self-supervised pre-training enables the network to pre-learn
how to estimate the optical flow from the generated data and
then guide it to learn optical flow estimation from the pose-
only training. We then introduce the details of the process.

\subsubsection{Training Sequence Generation}
To simulate image transformations caused by real motion, we employ the random homographic adaptation~\cite{superpoint, superglue} as shown in the left column of \cref{selftrain}. Furthermore, for a more faithful emulation of the diversities observed in genuine image sequences, some appearance augmentations are applied, including illumination, saturation, hue, and motion blur~\cite{superpoint}. Besides, the occlusion masks~\cite{selflow} based on superpixel, whose edges are often on the boundaries of the objects, are also added in appearance augmentations for better simulations of the occlusion problem in real-world scenarios. The transformation of the 2D points and the image after homographic transformations and appearance augmentation can be expressed as:
\begin{equation}
\label{Htrans}
\begin{aligned}
p^t = H^{t}p^0,\quad I^{t} = \mathcal{H}^t(A^t(I^0)),\quad t=1,2,..,N,
\end{aligned}
\end{equation}
where $H^t$ is the homograph matrix of the $t$-th frames;  $A^t(\cdot)$ and $\mathcal{H}^t$ are the appearance augmentation and the corresponding homograph transformation of the $t$-th image, respectively; $p^0$ and $p^t$ are the positions of pixels of the original frame and the $t$-th transformed frame, respectively; $I^0$ and $I^t$ are the original image and the $t$-th transformed image, respectively. Especially, since the spatial transformations tend to be more prominent with the index increasing, we apply $H^t$ with more considerable translation and rotation when the index $t$ increases. 

\subsubsection{Training Process}
As pointed out in~\cite{randomkp}, the features of the salient parts are instinctively different from the neighbors even though the features are extracted by an untrained network. Also, for better cooperation with the salient detection module, we use the strategy of using the patch set for tracking, which consists of both salient points and some random points. With this strategy, in the pre-training, the salient patches can be better distinguished from other points. As shown in \cref{selftrain}, the ground truth of point correspondance can be obtained by \cref{Htrans}. Since there are no 3D initial reprojection relationships, we assign the initial reprojection positions of the points around the target points with random displacements.

\subsubsection{Pre-Training Loss}
As pointed out in~\cite{rethinkingopticalflow}, the feature pre-training for improving the feature representations can improve the network ability for optical estimation. Therefore, the loss function is designed as two parts including feature loss, and flow loss, to make the network learn the feature better and optical estimation. 
We apply the softmax operation to get the probability of the feature correlation map,
\begin{equation}
\begin{aligned}
P(F^t, f^{0}_{l}) = {\rm softmax} (\gamma \cdot \frac{<F^{t}_{\hat{m_l}\hat{n_l}}, f^{0}_{l}>}{\sum_{m,n}  <F^{t}_{mn}, f^0_{l}>})
\end{aligned},
\end{equation}
where $f^{0}_{l}$ means the feature of the $k$-th points on the origin frame, $F_t$ means feature map of frame $t$, $F_{mn}^t$ denotes the $m$-th row and $n$-th column feature of $F^t$,  $F_{\hat{m_l}\hat{n_l}}^t$ denotes the feature that is corresponding with $f^0_{l}$ obtained by the homograpy transformation and $\gamma=10$ is the temperature parameter. Then, the cross-entropy loss as~\cite{s2dnet} is used to train the feature:
\begin{equation}
\begin{aligned}
L_{F} = - \frac{1}{\sum_t \sum_l O_{tl}} \sum_{t} \sum_{l} O_{tl} \log P(F^t,f^0_l)
\end{aligned},
\end{equation}
where $O_{tl}$ is the mask consisting of out-of-boundary mask and occlusion mask, i.e., if the $l$-th point on the $t$-th images is out-of-boundary or occluded, $O_{tl}=0$ and vice versa.
The flow estimation network with covariance can be trained as a negative log-likelihood:
\begin{small}
\begin{equation}
\begin{aligned}
L_{\delta} & = \sum_{(i,j,k)\in E} - \log P(\delta|I^t;\Psi) \\ 
      & =  \sum_{(i,j,k)\in E} (^{gt}\delta_{k}^{ij} - \delta_k^{ij})^{\scriptscriptstyle T} \Sigma^{ij}_{k} (^{gt}\delta_k^{ij} - \delta_{k}^{ij}) - \log(\det(\Sigma^{ij}_{k})), 
\end{aligned}
\end{equation}
\end{small}\noindent
where $^{gt}\delta_{k}^{ij}$, $\delta_{k}^{ij}$, and $\Sigma_{k}^{ij}$ are the ground truth flow obtained by the homography transformation, the estimated flow and the confidence weights of the $k$-th patch reprojected from $i$-th frame to $j$-th frame, respectively and the $(i,j,k) \in E$ means that the $k$-th patches on the $i$-th frame and the $i$-th frame and the $j$-th frame are defined as neighbor. The final loss is:
\begin{equation}
\begin{aligned}
L =  \lambda_{F_R} L_{F_R} + \lambda_{F_K} L_{F_K} + \lambda_{\delta} L_{\delta}
\end{aligned},
\end{equation}
where $L_{F_K}$ and $L_{F_R}$ denote the salient point features and random point features and empirically, the parameters are set as $\lambda_K = 1 > \lambda_R = 0.2$ and $\lambda_{\delta}=0.4$.

\subsection{Weighted Bundle Adjustment Layer}
After obtaining the sparse optical flow and the confident weights, the weighted bundle adjustment (BA) layer is applied to get the depths of the patches and the poses. It can be expressed by the weighted error of the estimated and the reprojected position as:
\begin{equation}
\begin{aligned}
T^*, d^* = \arg \min_{T,d} \sum_{(i,j,k)\in E} ||\hat{P}^{ij}_{k} - \tilde{P}^{ij}_{k}||^2_{\Sigma^{ij}_k},
\end{aligned}
\end{equation} 
where $\hat{P}^{ij}_{k}$, $\tilde{P}^{ij}_{k}$, and $\Sigma^{ij}_{k}$ denote the reprojected position, the estimated flow, and the confidence weights of the $k$-th patch from the $i$-th frame to the $j$-th frame, respectively. The Gauss-Newton iterative method is applied for optimization to obtain both the camera poses and the depths of the patches.

\subsection{Pose-Supervised Training}
After pre-training, the network is trained on the sequence images. The patches are randomly selected in this part. At the beginning of the training, translation normalization is used to eliminate the scale problem for better optical flow learning in the loss function. After the loss is not rapidly decreasing, we apply the Umeyama alignment algorithm~\cite{umeyama} to scale the estimated trajectory. The loss function can be shown as:
\begin{equation}\label{l2}
\begin{aligned}
L_{poses} = \sum_{ij} || Log_{SE(3)}[G^{-1}_{ij} T_{ij}]||_2,
\end{aligned}
\end{equation}
where $G_{ij}$ and $T_{ij}$ are the ground truth delta poses and the estimated delta poses between the $j$-th and the $i$-th frames. 

\subsection{Salient Patch Refining}
To make the network better cooperate with the salient patches, we then train it with a combination of the salient patches by the module described in \cref{selftrain} with the random patches. The set of the selected patches can be expressed as:
\begin{equation}
\begin{aligned}
P = P_K \cup P_R
\end{aligned},
\end{equation}
where $P_K$ denotes the random salient patches selected from the salient patch set and $P_R$ means the randomly selected patches. 
% We apply \cref{l2} for the pose supervision.
With this process, the systems can better use the salient patches for tracking and achieve better performance.

\begin{table}[t]
\caption{Performance comparisons on the TartanAir monocular test split on ATE(m). The $\bigtimes$ means tracking lost in the sequence.}
% \vspace{-5pt}
\centering
\label{tartanATE}
\begin{center}
\resizebox{1\columnwidth}{!}{
\begin{tabular}{c | c c c c c | c}
\toprule
Sequence & \makecell[c]{ORB-\\SLAM3\\~\cite{orbslam3}}& \makecell[c]{DSO\\~\cite{dso}} & \makecell[c]{COL-\\MAP\\~\cite{colmap}}  & \makecell[c]{DROID-\\VO\\~\cite{droidslam}} & \makecell[c]{DPVO\\~\cite{dpvo}} & Ours\\
\midrule
ME000 & 13.67  & 9.65   & 15.20  & 0.22 & \underline{0.13} & \textbf{0.11}\\ 
ME001 & 16.86  & 3.84   & 5.58  & 0.15 & \underline{0.14} & \textbf{0.06}\\
ME002 & 20.57  & 12.20 & 10.86 & \underline{0.24} & \textbf{0.13} & 0.29\\
ME003 & 16.00  & 8.17   & 3.93 & 1.27 & \underline{0.42} & \textbf{0.25} \\
ME004 & 22.27  & 9.27   & 2.62  & 1.04 & \underline{0.38} & \textbf{0.36} \\
ME005 & 9.28    & 2.94   & 14.78  & \underline{0.14}  & \underline{0.13} & \textbf{0.05}\\
ME006 & 21.61  & 8.15   & 7.00  &1.32 & \textbf{0.28} & \underline{0.48} \\
ME007 &  7.74   & 5.43   & 18.47 & 0.77 & \underline{0.14} & \textbf{0.11} \\
\midrule
MH000 & 15.44  & 9.92   & 12.26  & 0.32 & \underline{0.24} & \textbf{0.14} \\
MH001 & 2.92   & 0.35   & 13.45  & 0.13 & \underline{0.06} & \textbf{0.03}\\
MH002 & 13.51  & 7.96   & 13.45  & 0.08 & \textbf{0.05} & \textbf{0.05} \\
MH003 & 8.18   & 3.46   & 20.95  & 0.09 & \textbf{0.04} & \underline{0.06} \\
MH004 & 2.59   & $\bigtimes$ & 24.97  & 1.52 & \textbf{0.67} & \underline{1.23} \\
MH005 & 21.91  & 12.58  & 16.79  & 0.69 & \textbf{0.19} & \underline{0.23}\\
MH006 & 11.70  & 8.42   & 7.01   & 0.39 & \textbf{0.09} & \underline{0.17} \\
MH007 & 25.88  & 7.50   & 7.97   & 0.97 & \textbf{0.11} & \underline{0.22} \\
\midrule
Average & 14.38 & 7.32 & 12.50 & 0.33 & \textbf{0.20} & \underline{0.24}\\
\bottomrule
\end{tabular}}
\vspace{-0.2cm}
\end{center}
\end{table}

\begin{table}[t]
\caption{Performance comparisons on the TUM datasets on ATE(m). The $\bigtimes$ means tracking lost in the sequence.}
\vspace{-5pt}
\centering
\label{tumATE}
\begin{center}
\resizebox{1\columnwidth}{!}{
\begin{tabular}{c | c c c c c | c}
\toprule
Sequence &  \makecell[c]{ORB-\\SLAM3\\~\cite{orbslam3}} &  \makecell[c]{DSO\\~\cite{dso}} &  \makecell[c]{Tartan-\\VO~\cite{tartanvo}} &  \makecell[c]{DROID-\\VO~\cite{droidslam}} &  \makecell[c]{DPVO\\~\cite{dpvo}} & Ours\\
\midrule
360    & $\bigtimes$      & 0.173  & 0.178 & 0.161 & \underline{0.152} & \textbf{0.126}\\
desk   & \textbf{0.017}   & 0.567 & 0.125 & 0.028  & 0.028 & \underline{0.022}\\
desk2  & 0.210            & 0.916 & 0.122 & 0.099 & \underline{0.072} & \textbf{0.048}\\
floor  & $\bigtimes$      & 0.080 &  0.349 & \textbf{0.033} & 0.054 & \underline{0.047}\\
plant  & 0.034            & 0.121 & 0.297 & \textbf{0.028} & \underline{0.031} & 0.032\\
room   & $\bigtimes$      & 0.379 & 0.333 & \textbf{0.327} & 0.389 & \underline{0.357}\\
rpy    & $\bigtimes$      & 0.058 & 0.049 & \textbf{0.028} & \underline{0.037} & \textbf{0.028}\\
teddy  & $\bigtimes$      & $\bigtimes$ & 0.339 & 0.169 & \textbf{0.065} & \underline{0.097}\\
xyz    & \textbf{0.009}   & 0.031 &  0.062 & 0.013 & \underline{0.012} & 0.015\\
\midrule
Average & - & - & 0.206&0.098  & \underline{0.094} & \textbf{0.086}\\
\bottomrule
\end{tabular}
}
\vspace{-0.5cm}
\end{center}
\end{table}

\section{Experiments}

In this section, we perform experiments on popular benchmarks to quantitatively evaluate the effectiveness and the real-time capacity of our approach. Then, a live experiment is conducted in a highly challenging scenario to assess the robustness. Finally, some visual representations of selected patches, tracking outcomes, and correlation maps are presented to provide insights into the interpretability of our approach. 

\subsection{Implementation Details}

% This work is implemented by PyTorch~\cite{pytorch} and Lietorch~\cite{lietorch}. 
We conducted both training and evaluation using NVIDIA GPUs, with training performed on an NVIDIA RTX-3090 and evaluation on an NVIDIA RTX-4080. Our training procedure employed the AdamW optimizer and followed a one-cycle learning strategy. For pre-training, we randomly selected $120k$ images from the MegaDepth~\cite{megadepth} dataset and an additional $40k$ images from the NYU~\cite{nyuv2} dataset. This pre-training comprised two stages: feature pre-training (100k iterations, batch size 16, learning rate $1 \times 10^{-4}$) and flow pre-training (20k iterations, batch size 4, learning rate $5 \times 10^{-5}$), using a patch set consisting of $96$ salient patches and $32$ random patches. Subsequently, we trained the model on the TartanAir~\cite{tartanair} dataset for $240k$ steps, following DPVO~\cite{dpvo}, with only pose supervision on the pairs within $0 - 8.5$ average poses of the sequence and an initial learning rate of $8 \times 10^{-5}$. Lastly, we fine-tuned the salient patches refining module for $60k$ iterations, utilizing an initial learning rate of $2 \times 10^{-6}$ within $0 - 8$ average poses of the sequence, employing a patch set of $60$ salient patches and $20$ random patches, and selecting an NMS radius of $4$ during training. For evaluation, we set the NMS radius to $4$ and the removal window size to $18$. The patch size is set as $3$, i.e., the patch radius $r$ is $1$, and the number of patches of each frame is selected as $96$. The rest training and evaluation settings are the same as DPVO~\cite{dpvo}.

\begin{table}[t]
\caption{Performance comparisons on the EuRoC dataset on ATE(m). The $\bigtimes$ means tracking lost in the sequence.}
% \vspace{-5pt}
\centering
\label{eurocATE}
\begin{center}
\resizebox{1\columnwidth}{!}{
\begin{tabular}{c | c c c c c | c}
\toprule
Sequence &  \makecell[c]{SVO\\~\cite{svo}}&  \makecell[c]{DSO\\~\cite{dso}} &  \makecell[c]{Tartan-\\VO~\cite{tartanvo}} &  \makecell[c]{DROID\\-VO~\cite{droidslam}} &  \makecell[c]{DPVO\\~\cite{dpvo}} & Ours\\
\midrule
MH01 & 0.100     & \underline{0.046}  & 0.639 & 0.163 & 0.083 & \textbf{0.042} \\
MH02 & 0.120     & \underline{0.046}  & 0.325 & 0.121 & 0.052 & \textbf{0.045} \\
MH03 & 0.410     & 0.172 & 0.550 & 0.242 & \textbf{0.139} & \underline{0.151} \\
MH04 & 0.430     & 3.810 & 1.153 & 0.399 & \underline{0.172} & \textbf{0.134} \\
MH05 & 0.300    & 0.110 & 1.021 & 0.270 & \underline{0.116} & \textbf{0.091} \\
V101 & 0.070      & 0.089 & 0.447 & 0.103 & \underline{0.050} & \textbf{0.048} \\
V102 & 0.210      & 0.107 & 0.389 & \underline{0.165} & \textbf{0.168} & 0.183 \\
V103 & $\bigtimes$           & 0.903 & 0.622 & 0.158 & \underline{0.093} & \textbf{0.074} \\
V201 & 0.110      & \textbf{0.044} & 0.433 & 0.102 & 0.062 & \underline{0.053} \\
V202 & 0.110      & 0.132  & 0.749 & 0.115 & \underline{0.042} & \textbf{0.034} \\
V203 & 1.080      & 1.152 & 1.152 & \underline{0.204} & \textbf{0.125} & 0.207\\
\midrule
Average & - & 0.601 & 0.680 & 0.186  & \underline{0.100}& \textbf{0.096} \\
\bottomrule
\end{tabular}}
\vspace{-0.5cm}
\end{center}
\end{table}

\subsection{Quantitative Comparison}

We assess the estimated trajectory on ATE using the scale alignment with evo~\cite{evo} and compare with the state-of-the-art traditional and deep-learning-based methods on four public datasets: TartanAir~\cite{tartanair}, EuRoC~\cite{euroc}, TUM~\cite{tumrgbd}, and OIVIO~\cite{oivio} to show the efficiency of our method. Since some end-to-end methods, e.g., DeepVO~\cite{deepvo} and D3VO~\cite{d3vo}, are needed to retrain on new datasets, these methods are excluded for comparisons.
Following DPVO~\cite{dpvo}, we adopt DROID-VO as the VO version of DROID-SLAM~\cite{droidslam}. DPVO~\cite{dpvo} and DROID-VO are supervised by both optical flow and camera poses. We use the deterministic algorithm in both the network and the BA layer of our method for evaluation in this part.

\textbf{Performance Comparison}
We next evaluate the accuracy of our approach following the training and test split of TartanAir~\cite{tartanair}.
% Our approach is evaluated using the test split of TartanAir~\cite{tartanair} whose training split is our training dataset. 
As illustrated in \cref{tartanATE}, our method is compared against classic methods such as ORB-SLAM3~\cite{orbslam3}, DSO~\cite{dso}, and COLMAP~\cite{colmap}.
The performance of deep-learning-based methods such as DROID-VO~\cite{droidslam} and DPVO~\cite{dpvo} is also provided. Results of DPVO~\cite{dpvo} are reported as the mean results of 5 runs over all the experiments and use the default configuration. Our method can achieve that 9 of 16 sequences surpass the previous methods and have comparable average results compared with the best score. 

\textbf{Generalization Tests}
We employ the widely recognized EuRoC~\cite{euroc} and TUM~\cite{tumrgbd} datasets, which contain real-world collections that are extensively used for SLAM algorithm evaluations.
The results on EuRoC~\cite{euroc} are shown in \cref{eurocATE}, where we compare our method with the prior methods,~\cite{svo, dso, tartanvo, droidslam} and~\cite{dpvo}. Results of DPVO~\cite{dpvo} are reported as the mean results of 5 runs. Our method, only trained on TartanAir~\cite{tartanair}, attains the lowest ATE on most of the sequences (7 out of 11) and the lowest average ATE.
% compared with the other methods.

\cref{tumATE} shows the results of our method compared with the existing methods ORB-SLAM3~\cite{orbslam3}, DSO~\cite{dso}, TartanVO~\cite{tartanvo}, DROID-VO~\cite{droidslam} and DPVO~\cite{dpvo} on TUM~\cite{tumrgbd}. We resize the image into $320 \times 240$ and skip each frame as DROID-SLAM~\cite{droidslam} and DPVO~\cite{dpvo}. Since the image is downsampled, the NMS radius is chosen as $1$. Classic methods~\cite{orbslam3,dso} failed for sequences with a large movement, while deep-learning-based methods~\cite{droidslam, dpvo, tartanair} can handle. Our method has the lowest average trajectory error among these methods.

\begin{table}[t]
\caption{Performance comparisons on the OIVIO datasets on ATE(m). The $\bigtimes$ means tracking lost in the sequence.}
\vspace{-5pt}
\centering
\label{oiviocate}
\begin{center}
\resizebox{1\columnwidth}{!}{
\begin{tabular}{c |c c c c c c | c}
\toprule
Sequence &  \makecell[c]{ORB-\\SLAM3\\~\cite{orbslam3}} & \makecell[c]{SVO\\~\cite{svo}} & \makecell[c]{DSO\\~\cite{dso}} & \makecell[c]{Tartan-\\VO~\cite{tartanvo}} & \makecell[c]{DROID\\-VO~\cite{droidslam}} & \makecell[c]{DPVO\\~\cite{dpvo}}& Ours \\
\midrule
MN\_015\_GV\_01 & 0.178 & $\bigtimes$ & 1.615& 1.717 & \underline{0.222} & \textbf{0.211} & 0.307\\
MN\_050\_GV\_01 &  0.222 & $\bigtimes$ & 0.721& 1.400 & \underline{0.165} & 0.249 & \textbf{0.135}\\
MN\_100\_GV\_01 & 0.171&  3.488 & 1.182& 1.357 & \underline{0.139} & 0.164 & \textbf{0.117}\\
MN\_015\_GV\_02 & 0.119& 3.384 & 1.427& 2.270 & 0.115 & \underline{0.089} & \textbf{0.063}\\
MN\_050\_GV\_02 & 0.104 & 2.661 & 0.598& 1.716 & 0.092 & \underline{0.088} & \textbf{0.076}\\
MN\_100\_GV\_02 & 0.096 & 3.245 & 0.263& 1.717 & 0.105 & \underline{0.091} & \textbf{0.076}\\
TN\_015\_GV\_01 & 0.323 & $\bigtimes$ & 0.223& 2.522 & \textbf{0.051} & 0.112 & \underline{0.110}\\
TN\_050\_GV\_01 & 0.269 & $\bigtimes$ & 0.325& 2.424 & 0.236 & \underline{0.133} & \textbf{0.094}\\
TN\_100\_GV\_01 & 0.255 & $\bigtimes$ & 0.512& 2.888 & \underline{0.125} & 0.146 & \textbf{0.100}\\
\midrule
Average & 0.193 & - & 0.763 & 2.00 & \underline{0.139} & 0.143 & \textbf{0.120} \\
\bottomrule
\end{tabular}}
\vspace{-0.2cm}
\end{center}
\end{table}

\begin{table}[t]
\caption{Comparsions of average runtime.}
\vspace{-5pt}
\centering
\label{TimeEval}
\begin{center}
\begin{tabular}{c | c c}
\toprule
Metric & DPVO~\cite{dpvo} & Ours \\
\midrule
% Feature extraction & 8.14& 4.73\\
% Flow update & 22.89 & 22.78\\
% BA & 1.17 & 13.41\\
Total time (ms) & 37.2 & 41.3\\
% GPU Memory (GB) &  2.1 & 3.7\\
% std (m) & 0.048 & 0.012 \\
\bottomrule
\end{tabular}
\label{timeanalysis}
\end{center}
\vspace{-0.6cm}
\end{table}

\textbf{Illumination Test} 
We evaluated our method's robustness against darkness and illumination changes compared to SOTA methods, detailed in \cref{oiviocate} using the OIVIO dataset, which is captured with onboard lighting in dark environments. ORB-SLAM3 results are taken from~\cite{afeorbslam}. Traditional methods, especially direct ones~\cite{svo,dso}, perform poorly under changing illuminations. Our approach shows a $13.6\%$ improvement in average ATE over the best previous result.

% To demonstrate the robustness of our approach in environments characterized by darkness and changing illuminations, we compare our method with existing methods in \cref{oiviocate}, focusing on the OIVIO~\cite{oivio} which is gathered with an onboard light in dark channels. The results of ORB-SLAM3~\cite{orbslam3} are derived from~\cite{afeorbslam}.  The changing illuminations degenerate the performances of traditional methods, especially direct approaches~\cite{svo,dso}, which show bad performance compared with deep learning-based methods. 
% In general, learning-based methods are more robust to illumination changes and our approach has a $13.6\%$ improvement over the best former result on the average ATE.
\subsection{Runtime Analysis}
In \cref{TimeEval}, we compare our method with DPVO~\cite{dpvo} with default configurations in terms of the average runtime. In experiments, we found that DPVO~\cite{dpvo} uses non-deterministic CUDA operations in the BA layer, which are faster but often produce large performance fluctuations due to numerical instability. To solve this problem, we use the nondeterministic operations in the BA layer and NVIDIA TensorRT Toolkit for acceleration. 
Nevertheless, we can still achieve real-time inference but with stabler performance in terms of accuracy.

\subsection{Qualitative Experiment}
In this part, we first compare our patch selection strategy between our method and DPVO~\cite{dpvo} to show the necessity of selecting salient patches in extreme situations. Then, a live experiment with enormous illumination changes is conducted to show the robustness of our method in demanding practical scenarios. Finally, we visualize the correlation map and the flow estimation of the three different types of points to show the importance of the feature representations and the salient points selection strategy in demanding scenarios.

\textbf{Patches Selection}
We compared the patches selection strategy and our method in \cref{kpcompare}. DPVO~\cite{dpvo} randomly selects the patches for tracking, and our strategy selects the salient patches. We can see that the patches selected by DPVO are likely to be uneven and located in textureless (invalid) places. If the scene is very rich in texture, the random selection can still work well since random selection can select a sufficient number of effective patches and estimate the flow well with dense flow supervision. However, when dealing with textureless scenes, the strategic selection of salient patches ensures a reliable tracking of informative regions, thereby significantly enhancing both the robustness and accuracy.
% within such challenging environments.

\begin{figure}[t]
\centering 
\includegraphics[width=\linewidth]{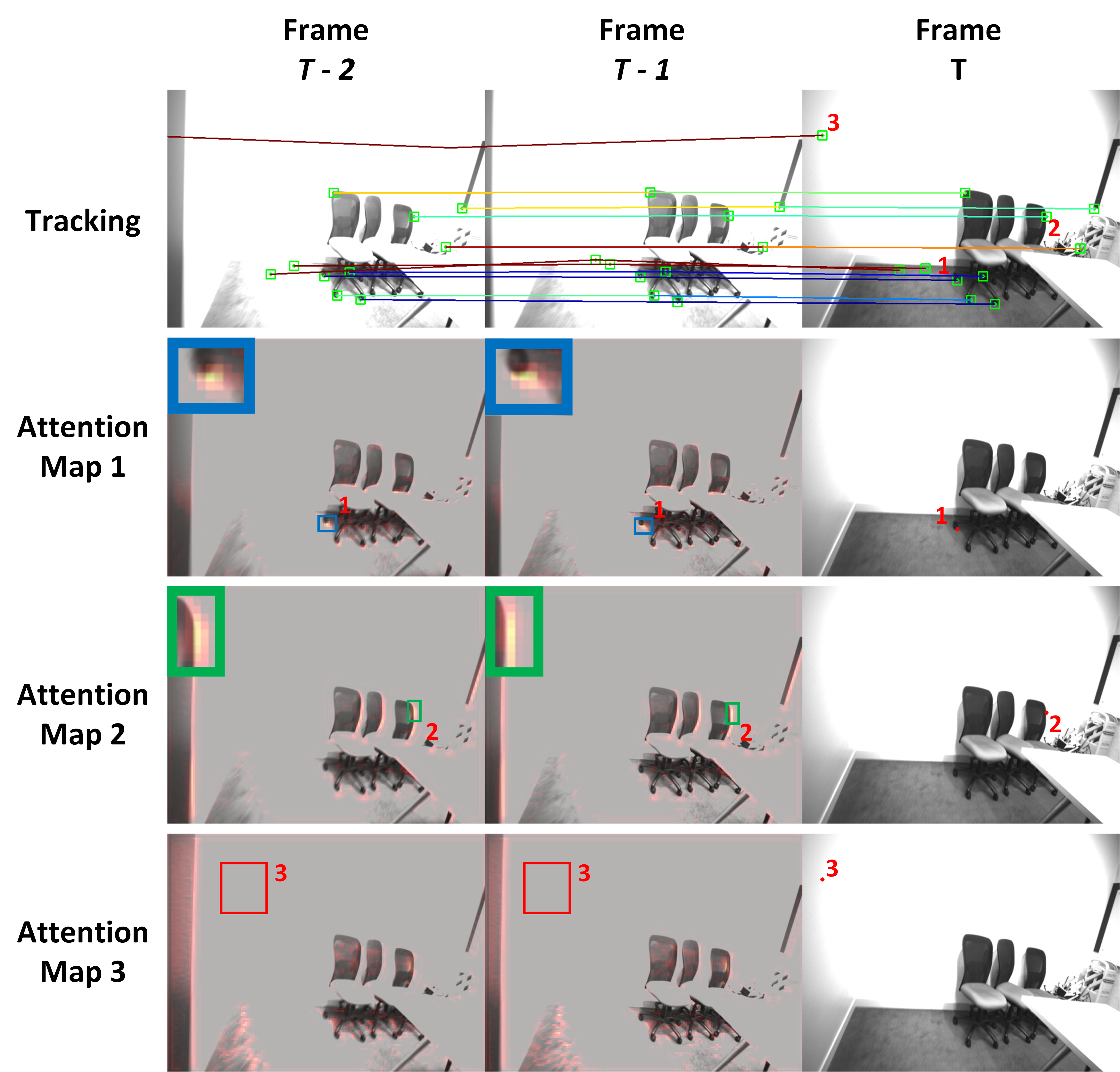}
\caption{The visualizations of the patches tracking and the corresponding confidence weights, and the correlation map. The first row displays tracked patches alongside their corresponding confidence weights (bluer tones signify higher scores, whereas redder tones denote lower scores). Rows two through four depict the attention map, with darker shades representing lower attention scores while lighter shades signifying higher scores. The magnified content from the square-boxed area can be found in the top-left corner.}
\vspace{-0.5cm}
\label{fig:heatmap}
\end{figure}

\begin{figure*}[t]
\centering 
\includegraphics[width=0.9\linewidth]{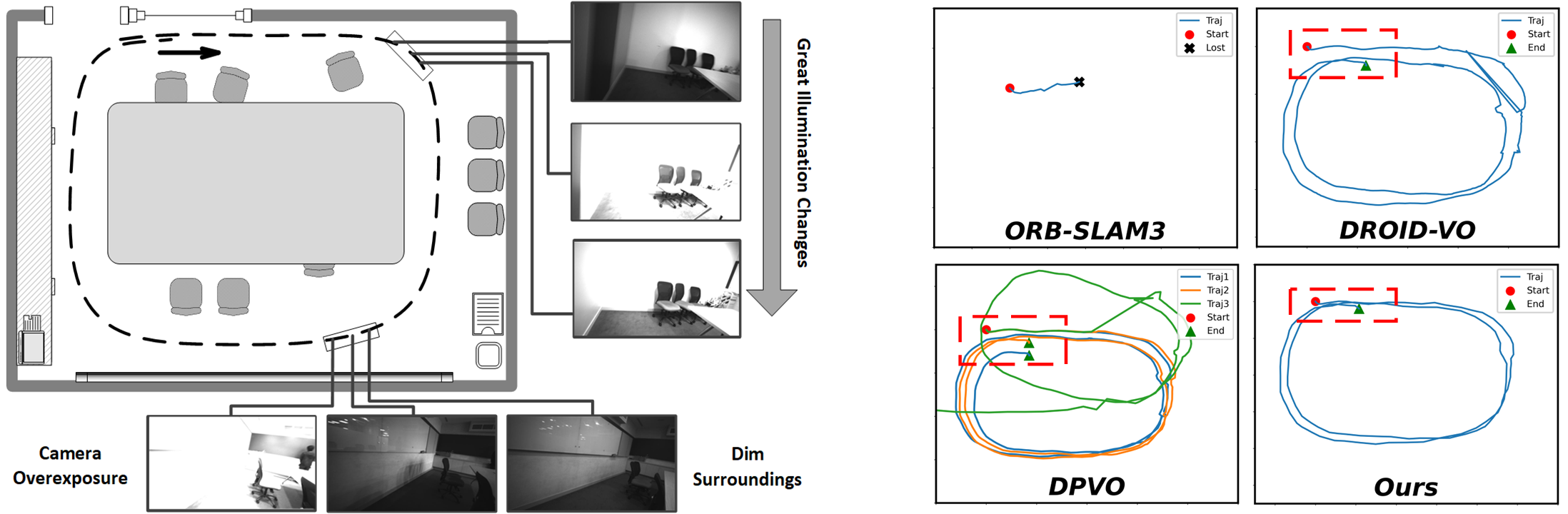}
\caption{A real-world dome in a meeting room with significant illumination changing to compare the generalization ability of different methods. The approximate trajectory involved walking around the table in a nearly identical path for two loops and the initial path and the end path are roughly aligned. For lack of the ground truth of the sequence, we present the trajectory in four images and assess the performances based on the degree of overlapping of the trajectory of the initial and final phases.}
\label{fig::meetingroom} 
% \vspace{-5pt}
\end{figure*}

\textbf{Correlation and Tracking Relationship Analysis} To demonstrate the importance of pre-training and the necessity of the salient patches selection, we present the tracking result and demonstrate the relationship between the features and flow estimation in \cref{fig:heatmap}.
From its first row, we can see that the textureless parts have low confidence weight while the rich information parts like the edge of the chair backrests and legs have higher weights, which represents that the salient patches are more likely to give effective tracking for pose estimation. 
We also compare the attention map of three kinds of points to show the relationship between the tracking quality and the attention map (feature) from the second row to the fourth row.
It can be observed that the attention maps exhibit varying confidence weights: higher attention confidence maps (Map 1) produce better tracking quality than lower confidence maps (Map 3).
% confidence maps are concentrated (Attention Map 1), while middle-confidence maps are dispersed (Attention Map 2), and low-confidence maps are very low (Attention Map 3).
% It can be observed that the attention map of high confidence weight is high and concentrated as shown in Attention Map 1.
% The attention map of middle confidence weight is relatively low and dispersed as shown in Attention Map 2.
% The attention map of low confidence weight is very low as shown in the Attention Map 3.
This means that the tracking quality is highly related to the correlation maps. With better feature representations and patch selection, the tracking quality can be improved.

\begin{table}[t]
\caption{Ablation studies on the main designs of our method. HPT means the homography pre-training. SPR means the salient patches refining. The table shows the average ATE of the datasets.}
\centering
\vspace{-0.2cm}
\label{AblationStudies}
\begin{center}
\resizebox{\columnwidth}{!}{
\begin{tabular}{c | c c c c}
\toprule
Datasets & TartanAir & EuRoC & TUM & OIVIO \\
\midrule
DROID-VO &  0.58 & 0.186 & 0.098 & 0.139\\
DPVO & 0.20  & 0.100   & 0.094  & 0.143\\
% DVPO (Salient Patch) & \textbf{0.19} & 0.141& 0.095 & 0.131 \\
\midrule
w/o HPT  w/o SPR & 0.333   & 0.213  &  0.116  &0.136\\
w/o HPT &0.30&0.20& 0.099 & \textbf{0.11} \\
w/o SPR & 0.316 & 0.118  &  0.099 &  0.126 \\ 
\midrule
Full & \textbf{0.24}   & \textbf{0.096}  &  \textbf{0.086} & \underline{0.120}\\
\bottomrule
\end{tabular}}
\end{center}
\end{table}

\begin{table}[t]
\caption{Ablation studies on salient patch refining strategies.}
\vspace{-0.2cm}
\centering
\label{spr}
\begin{center}
\resizebox{\columnwidth}{!}{
\begin{tabular}{c | c c c c}
\toprule
Datasets & TartanAir & EuRoC & TUM & OIVIO \\
\midrule
Salient Patches Only & 0.27 & 0.099 &  0.108 & 0.138 \\
Combined Salient and Random Patches (Adopted) & \textbf{0.24}   & \textbf{0.096}  &  \textbf{0.086} & \textbf{0.120}\\
\bottomrule
\end{tabular}}
\vspace{-0.8cm}
\end{center}
\end{table}

\subsection{Real-world Robustness and Generalization Test} 

To demonstrate the robustness and generalization ability of our method in the challenging scenario with significant photometric changes, we test our method in a live sequence in a meeting room collected by an Intel RealSense. 
Specifically, the lighting in a meeting room is regularly switched on and off, causing abrupt fluctuations in illumination and frequent camera overexposure.
For easier comparison, we make sure the initial and final locations on paths are kept roughly the same.
These common yet hard scenarios cause significant challenges for visual odometry. The meeting room layout, some sampled images, and the rough trajectory are presented in \cref{fig::meetingroom}. 

We compared the trajectory of our method with ORB-SLAM3~\cite{orbslam3}, DROID-VO~\cite{droidslam}, and DPVO~\cite{dpvo} on the sequence in  \cref{fig::meetingroom}. ORB-SLAM3~\cite{orbslam3} loses track in the sequence. DROID-VO~\cite{droidslam} shows a significant drift between the end path and the start path. Since DPVO~\cite{dpvo} cannot have a fixed performance, we run DPVO three times. We can see that in the three runs, one fails and the other two runs have massive drifts between the end and start location. Our methods show obvious consistency in the end and start locations. This demonstrates that our method is much more robust and has better generalization ability than prior works. 

\begin{table}[t]
\caption{Ablation studies on homography pretraining strategies.}
\centering
\vspace{-0.2cm}
\label{Ablationkp}
\begin{center}
\resizebox{1\columnwidth}{!}{
\begin{tabular}{c | c c c c}
\toprule
Datasets & TartanAir~\cite{tartanair} & EuRoC~\cite{euroc} & TUM~\cite{tumrgbd} & OIVIO~\cite{oivio} \\
\midrule
% Baseline &0.30&0.20& 0.99 & \textbf{0.11} \\
% \midrule
Random Patch Set & 0.25   & 0.183  &  \textbf{0.070} &0.133\\
% KP-Pretrain & 0.24 &0.096  &  0.086 &  0.120 \\ 
Without Flow & \textbf{0.23} & 0.18 & \underline{0.079} & \underline{0.127} \\
\midrule
\makecell[c]{SP Combining Set (Adopted)} & \underline{0.24} & \textbf{0.096}  & 0.086 &  \textbf{0.120} \\ 
\bottomrule
\end{tabular}}
\vspace{-0.6cm}
\end{center}
\end{table}

\subsection{Ablation Studies}
In this section, we analyze the influence of our primary design elements: pre-training and salient patch refinement. We also study the impacts of the different designs of the pre-training strategy on the four former used datasets.
We show the experimental results of the proposed components of our method, the homography pre-training, and the salient patch refining in \cref{AblationStudies}.
We can see that when the system is trained without the pre-training part and the salient patch refining module, the model shows the worst results on the 4 datasets.
When the pre-training module is enabled, the results of the 4 datasets are greatly improved, and the highest and minimum increase is $44.6\%$ and $5.2\%$ compared with the original model, respectively.
When the salient part refining module is enabled, the results are improved on 3 datasets, apart from OIVIO, from $0.06\%$ to $19\%$. 
% In \Cref{AblationStudies}, we also compare DPVO directly added with salient patches 

In \Cref{spr}, we also compare the salient patches only strategy with the combined salient and random patches strategy in the salient patch refining. The latter strategy can achieve better results since it can simultaneously increase the network's cooperation with the salient patches and introduce more diverse training data distributions.
% The full system can show the best results on OIVIO, which contains a large number of small motions, making it easier for learning-based methods to fit the pose labels.

In \cref{Ablationkp}, we compare three designs of homography pre-training: Random Patch Set, Without Flow and SP Combining Set. The Random Patch Set means that the patch set is selected randomly in pre-training. The Without Flow means that the pre-training model is trained without flow loss in pre-training. The SP Combining Set means the patch set consists of the salient patches and the random patches in pre-training. 
% Compared with the model without pre-training as shown in the \cref{AblationStudies}, all of the pretraining designs can generally improve the results. 
We can see that the SP Combining Set that we adopt can achieve the most balanced results compared to other designs.

\section{Conclusions}
Our study introduces a sparse hybrid visual odometry method without requiring labeled optical flow data. We utilize homographic self-supervised pre-training to teach the flow estimation network motion information, benefiting downstream tasks training like visual odometry with pose-only supervision. To increase robustness and accuracy in real-world scenarios, we introduce salient patch selection and refining modules. Our experiments, on four public datasets and a live demo, showcase strong generalization, robustness, and accuracy. 
One limitation of the proposed method is its weak robustness in the situations with an extreme high number of dynamic objects, which can be further improved in future works.

% \section*{Acknowledgement}
% \textbf{Acknowledgement} 
% This work is supported by NTU Presidential Postdoctoral Fellowship.
\bibliographystyle{ieeetr}
\bibliography{ref.bib}

\end{document}